\documentclass[11pt,a4paper]{article}
\usepackage[hyperref]{acl2019}
\usepackage{times}
\usepackage{latexsym}
\usepackage{url}
\usepackage{paralist}
\usepackage{multicol,graphicx,mathrsfs,pifont,amscd,latexsym,color,url,longtable, pifont, subfigure,epstopdf,setspace,colortbl, booktabs, bbm,listings, balance,color,indentfirst,algorithmic,algorithm,enumitem, subfigure,amsmath,amssymb, verbatim}

\usepackage{tabularx, makecell, multirow} 

\aclfinalcopy 
\def\aclpaperid{2405} 

\newcommand{\YS}[2][]{\note[#1]{YS}{green}{#2}}

\newcommand{\TC}[2][]{\note[#1]{TC}{red}{#2}}

\usepackage{todonotes}
\makeatletter
\newcommand*\iftodonotes{\if@todonotes@disabled\expandafter\@secondoftwo\else\expandafter\@firstoftwo\fi}  
\makeatother

\newcommand{\note}[4][]{\todo[author=#2,color=#3,size=\scriptsize,fancyline,caption={},#1]{#4}} 

\newcommand{\ting}[1]{\textcolor{blue}{[\textit{TC: #1}]}} 

\title{Few-Shot Representation Learning for Out-Of-Vocabulary Words}

\author{Ziniu Hu, Ting Chen, Kai-Wei Chang, Yizhou Sun\\
  University of California, Los Angeles\\
  \texttt{\{bull, tingchen, kwchang, yzsun\}@cs.ucla.edu}
}

\date{}
\begin{document}
\maketitle
\begin{abstract}
Existing approaches for learning word embeddings often assume there are sufficient occurrences for each word in the corpus, such that the representation of words can be accurately estimated from their contexts. However, in real-world scenarios, out-of-vocabulary  (a.k.a. OOV) words that do not appear in training corpus emerge frequently. It is challenging to learn accurate representations of these words with only a few observations. In this paper, we formulate the learning of OOV embeddings as a few-shot regression problem, and address it by training a representation function to predict the oracle embedding vector (defined as embedding trained with abundant observations) based on limited observations. Specifically, we propose a novel hierarchical attention-based architecture to serve as the neural regression function, with which the context information of a word is encoded and aggregated from K observations. Furthermore, our approach can leverage Model-Agnostic Meta-Learning (MAML) for adapting the learned model to the new corpus fast and robustly. Experiments show that the proposed approach significantly outperforms existing methods in constructing accurate embeddings for OOV words, and improves downstream tasks where these embeddings are utilized.

\end{abstract}
\section{Introduction}\label{sec:introduction}
Distributional word embedding models aim to assign each word a low-dimensional vector representing its semantic meaning. These embedding models have been used as key components in natural language processing systems. To learn such embeddings, existing approaches such as skip-gram
models~\cite{DBLP:conf/nips/MikolovSCCD13} resort to an auxiliary task of predicting the context words (words surround the target word). These embeddings have shown to be able to capture syntactic and semantic relations between words. 

Despite the success, an essential issue arises: most existing embedding techniques assume the availability of abundant observations of each word in the training corpus. 
When a word occurs only a few times during training (i.e., in the few-shot setting), the corresponding embedding vector is not accurate~\cite{DBLP:conf/eacl/CohnBNAM17}. In the extreme case, some words are not observed when training the embedding, which are known as out-of-vocabulary (OOV) words. These words are often rare and might only occurred for a few times in the testing corpus. Therefore, the insufficient observations hinder the existing context-based word embedding models to infer accurate OOV embeddings. This leads us to the following research problem: {\it How can we learn accurate embedding vectors for OOV words during the inference time by observing their usages for only a few times? }

Existing approaches for dealing with OOV words can be categorized into two groups. 
The first group of methods derives embedding vectors of OOV words based on their morphological information \cite{bojanowski2016enriching, DBLP:conf/aaai/KimJSR16,DBLP:conf/emnlp/PinterGE17}. This type of approaches has a limitation when the meaning of words cannot be inferred from its subunits (e.g., names, such as Vladimir).
The second group of approaches attempts to learn to embed an OOV word from a few examples. In a prior study~\cite{DBLP:conf/eacl/CohnBNAM17, DBLP:conf/emnlp/HerbelotB17}, these demonstrating examples are treated as a small corpus and are used to fine-tune OOV embeddings. Unfortunately, fine-tuning with just a few examples usually leads to overfitting. In another work~\cite{DBLP:conf/acl/AroraLMKSS18}, a simple linear function is used to infer embedding of an OOV word by aggregating embeddings of its context words in the examples. However, the simple linear averaging can fail to capture the complex semantics and relationships of an OOV word from its contexts.

Unlike the existing approaches mentioned above, humans have the ability to infer the meaning of a word based on a more comprehensive understanding of its contexts and morphology.
Given an OOV word with a few example sentences, humans are capable of understanding the semantics of each sentence, and then aggregating multiple sentences to estimate the meaning of this word.
In addition, humans can combine the context information with sub-word or other morphological forms to have a better estimation of the target word.
Inspired by this, we propose an attention-based hierarchical context encoder (HiCE), which can leverage both sentence examples and morphological information. Specifically, the proposed model adopts multi-head self-attention to integrate information extracted from multiple contexts, and the morphological information can be easily integrated through a character-level CNN encoder.

In order to train HiCE to effectively predict the embedding of an OOV word from just a few examples, we introduce an episode based few-shot learning framework. In each episode, we suppose a word with abundant observations is actually an OOV word, and we use the embedding trained with these observations as its oracle embedding. Then, the HiCE model is asked to predict the word's oracle embedding using only the word's $K$ randomly sampled observations as well as its morphological information. This training scheme can simulate the real scenarios where OOV words occur during inference, while in our case we have access to their oracle embeddings as the learning target. Furthermore, OOV words may occur in a new corpus whose domain or linguistic usages are different from the main training corpus. To deal with this issue, we propose to adopt Model-Agnostic Meta-Learning (MAML)~\cite{DBLP:conf/icml/FinnAL17} to assist the fast and robust adaptation of a pre-trained HiCE model, which allows HiCE to better infer the embeddings of OOV words in a new domain by starting from a promising initialization.

We conduct comprehensive experiments based on both intrinsic and extrinsic embedding evaluation. Experiments of intrinsic evaluation on the Chimera benchmark dataset demonstrate that the proposed method, HiCE, can effectively utilize context information and outperform baseline algorithms. For example,  HiCE achieves $9.3\%$ relative improvement in terms of Spearman correlation compared to the state-of-the-art approach, $\grave{a} \ la\ carte$, regarding 6-shot learning case.
Furthermore, with experiments on extrinsic evaluation, we show that our proposed method can benefit downstream tasks, such as named entity recognition and part-of-speech tagging, and outperform existing methods significantly.

The contributions of this work can be summarized as follows.
\begin{compactitem}
\item We formulate the OOV word embedding learning as a $K$-shot regression problem and propose a simulated episode-based training schema to predict oracle embeddings.
\item We propose an attention-based hierarchical context encoder (HiCE) to encode and aggregate both context and sub-word information. We further incorporate MAML for fast adapting the learned model to the new corpus by bridging the semantic gap.
\item We conduct experiments on multiple tasks, and through quantitative and qualitative analysis, we demonstrate the effectiveness of the proposed method in fast representation learning of OOV words for down-stream tasks.
\end{compactitem}
\section{The Approach}\label{sec:approach}
In this section, we first formalize the problem of OOV embedding learning as a few-shot regression problem. Then, we present our embedding prediction model, a hierarchical context encoder (HiCE) for capturing the semantics of context as well as morphological features. Finally, we adopt a state-of-the-art meta-learning algorithm, MAML, for fast and robust adaptation to a new corpus.

\subsection{The Few-Shot Regression Framework}

\paragraph{Problem formulation} We consider a training corpus $D_T$, and a given word embedding learning algorithm (e.g., Word2Vec) that yields a learned word embedding for each word $w$, denoted as $T_w\in \mathbb{R}^{d}$. 
Our goal is to infer embeddings for OOV words that are not observed in the training corpus $D_T$ based on a new testing corpus $D_N$. 

$D_N$ is usually much smaller than $D_T$ and the OOV words might only occur for a few times in $D_N$, thus it is difficult to directly learn their embedding from $D_N$. Our solution is to learn an neural regression function $F_{\theta}(\cdot)$ parameterized with $\theta$ on $D_T$. The function $F_{\theta}(\cdot)$ takes both the few contexts and morphological features of an OOV word as input, and outputs its approximate embedding vector. The output embedding is expected to be close to its \textit{``oracle'' embeddings} vector that assumed to be learned with plenty of observations. 

To mimic the real scenarios of handling OOV words, we formalize the training of this model in a few-shot regression framework, where the model is asked to predict OOV word embedding with just a few examples demonstrating its usage. The neural regression function $F_{\theta}(\cdot)$ is trained on $D_T$, where we pick $N$ words $\{w_t\}_{t=1}^N$ with sufficient observations as the target words, and use their embeddings $\{T_{w_t}\}_{t=1}^N$ as \textit{oracle embeddings}.  For each target word $w_t$, we denote $S_t$ as all the sentences in $D_T$ containing $w_t$. 
It is worth noting that we exclude words with insufficient observations from target words due to the potential noisy estimation for these words in the first place.

In order to train the neural regression function $F_{\theta}(\cdot)$, we form episodes of few-shot learning tasks. In each episode, we randomly sample $K$ sentences from $S_t$, and mask out $w_t$ in these sentences to construct a masked supporting context set $\mathbf{S}_t^K = \{s_{t,k}\}_{k=1}^K$, 
where $s_{t,k}$ denotes the $k$-th masked sentence for target word $w_t$. We utilize its character sequence as features, which are denoted as $C_t$.
Based on these two types of features, the model ${F_{\theta}}$ is learned to predict the oracle embedding. In this paper, we choose cosine similarity as the proximity metric, due to its popularity as an indicator for the semantic similarity between word vectors. The training objective is as follows.
\begin{align}
\hat{\theta} \!=\! \mathop{\arg\max_\theta} \sum_{\substack{w_t}}\sum_{\substack{\mathbf{S}_t^K \sim \mathbf{S}_t}}\! \cos\left(F_{\theta}(\mathbf{S}_t^K, C_t), T_{w_t}\right), \label{objective}
\end{align}
where $\mathbf{S}_t^K \sim \mathbf{S}_t$ means that the $K$ sentences containing target word $w_t$ are randomly sampled from all the sentences containing $w_t$. Once the model ${F_{\hat{\theta}}}$ is trained (based on $D_T$), it can be used to predict embedding of OOV words in $D_N$ by taking all sentences containing these OOV words and their character sequences as input.

\begin{figure}
\begin{center}
\includegraphics[width=1.0\columnwidth]{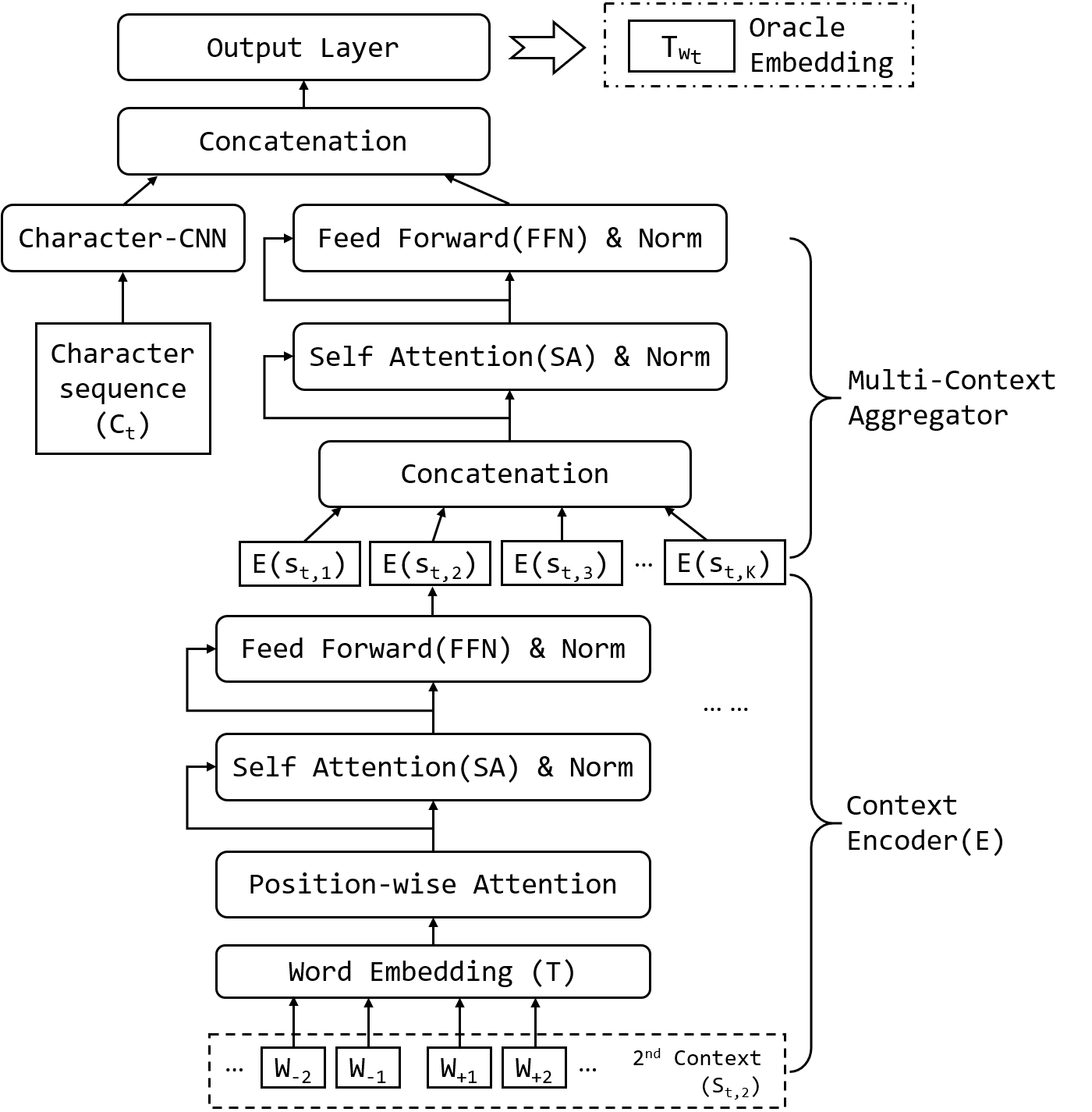}
\caption{The proposed hierarchical context encoding architecture (HiCE) for learning embedding representation for OOV words.}\label{fig:encoder}
\end{center}
\end{figure}

\subsection{Hierarchical Context Encoding (HiCE)}

Here we detail the design of the neural regression function $F_{\theta}(\cdot)$. Based on the previous discussion, $F_{\theta}(\cdot)$ should be able to analyze the complex semantics of context, to aggregate multiple pieces of context information for comprehensive embedding prediction, and to incorporate morphological features. These three requirements cannot be fulfilled using simple models such as linear aggregation~\cite{DBLP:conf/acl/AroraLMKSS18}.

Recent progress in contextualized word representation learning~\cite{DBLP:conf/naacl/PetersNIGCLZ18,bert} has shown that it is possible to learn a deep model to capture richer language-specific semantics and syntactic knowledge purely based on self-supervised objectives. Motivated by their results, we propose a hierarchical context encoding (HiCE) architecture to extract and aggregate information from contexts, and morphological features can be easily incorporated. Using HiCE as $F_{\theta}(\cdot)$, a more sophisticated model to process and aggregate contexts and morphology can be learned to infer OOV embeddings.

\paragraph{Self-Attention Encoding Block}

Our proposed HiCE is mainly based on the self-attention encoding block proposed by~\citet{DBLP:conf/nips/VaswaniSPUJGKP17}. Each encoding block consists of a self-attention layer and a point-wise, fully connected layer. Such an encoding block can enrich the interaction of the sequence input and effectively extract both local and global information.

Self-attention (SA) is a variant of attention mechanism that can attend on a sequence by itself. 
For each head $i$ of the attention output, we first transform the sequence input matrix $x$ into query, key and value matrices, by a set of three different linear projections $W_i^Q, W_i^K, W_i^V$. Next we calculate matrix product $xW_i^Q (xW_i^K)^T$,
then scale it by the square root of the dimension of the sequence input $\frac{1}{\sqrt{d_{x}}}$ to get mutual attention matrix of the sequence. Finally we aggregate the value matrices using the calculated attention matrix, and get $a_{self,i}$ as the self attention vector for head $i$:
$$
    a_{self,i} = {\rm softmax}\left(\frac{xW_i^Q \ (xW_i^K)^T}{\sqrt{d_{x}}}\right)xW_i^V.
$$

Combining all these self-attentions $\{a_{self,i}\}_{i=1}^h$ by a linear projection $W^O$, we have a $SA(x)$ with totally $h$ heads, which can represent different aspects of mutual relationships of the sequence $x$:
$$
    SA(x) = {\rm Concat}(a_{self,1},..., a_{self,h})W^O.
$$

The self-attention layer is followed by a fully connected feed-forward network (FFN), which applies a non-linear transformation to each position of the sequence input $x$. 

For both SA and FFN,
we apply residual connection~\cite{DBLP:conf/cvpr/HeZRS16} followed by layer normalization~\cite{DBLP:journals/corr/BaKH16}.
Such a design can help the overall model to achieve faster convergence and better generalization.

In addition, it is necessary to incorporate position information for a sequence. Although it is feasible to encode such information using positional encoding, our experiments have shown that this will lead to bad performance in our case. Therefore, we adopt a more straightforward position-wise attention, by multiplying the representation at $pos$ by a positional attention digit $a_{pos}$. In this way, the model can distinguish the importance of different relative locations in a sequence.

\paragraph{HiCE Architecture}

As illustrated in Figure \ref{fig:encoder}, HiCE consists of two major layers: the \textit{Context Encoder} and the \textit{Multi-Context Aggregator}. 

For each given word $w_t$ and its $K$ masked supporting context set $\mathbf{S}_t^K$ = $\{s_{t,1}, s_{t,2}, ..., s_{t,K}\}$, 
a lower-level \textit{Context Encoder} ($E$) takes each sentence $s_{t,k}$ as input, followed by position-wise attention and a self-attention encoding block, and outputs an encoded context embedding $E(s_{t,k})$. On top of it, a \textit{Multi-Context Aggregator} combines multiple encoded contexts, i.e., $E(s_{t,1}), E(s_{t,2}), ..., E(s_{t,K})$, by another self-attention encoding block. Note that the order of contexts can be arbitrary and should not influence the aggregation, we thus do not apply the position-wise attention in \textit{Multi-Context Aggregator}. 

Furthermore, the morphological features can be encoded using character-level CNN following \cite{DBLP:conf/aaai/KimJSR16}, which can be concatenated with the output of \textit{Multi-Context Aggregator}. Thus, our model can leverage both the contexts and morphological information to infer OOV embeddings.


\subsection{Fast and Robust Adaptation with MAML}

So far, we directly apply the learned neural regression function ${F_{\hat{\theta}}}$ trained on $D_T$ to OOV words in $D_N$. This can be problematic when there exists some linguistic and semantic gap between $D_T$ and $D_N$. For example, words with the same form but in different domains~\cite{DBLP:conf/acl/SarmaLS18} or at different times~\cite{DBLP:conf/acl/HamiltonLJ16} can have different semantic meanings. Therefore, to further improve the performance, we aim to adapt the learned neural regression function ${F_{\hat{\theta}}}$ from $D_T$  to the new corpus $D_N$. 
A na\"ive way to do so is directly fine-tuning the model on $D_N$. However, in most cases, the new corpus $D_N$ does not have enough data compared to $D_T$, and thus directly fine-tuning on insufficient data can be sub-optimal and prone to overfitting.

To address this issue, we adopt Model Agnostic Meta-Learning (MAML)~\cite{DBLP:conf/icml/FinnAL17} to achieve fast and robust adaption. 
Instead of simply fine-tuning ${F_{\hat{\theta}}}$ on $D_N$, 
MAML provides a way of learning to fine-tune. That is, the model is firstly trained on $D_T$ to get a more promising initialization, based on which fine-tuning the model on $D_N$ with just a few examples could generalize well.

More specifically, in each training episode, we first conduct gradient descent using sufficient data in $D_T$ to learn an updated weight $\theta^*$. For simplification, we use $\mathcal{L}$ to denote the loss function of our objective function (\ref{objective}). The update process is as:
\begin{equation*}
\theta^* = \theta - \alpha \nabla_\theta \mathcal{L}_{D_T}(\theta).
\end{equation*}
We then treat $\theta^*$ as an initialized weight to optimize $\theta$ on the limited data in $D_N$. The final update in each training episode can be written as follows.
\begin{align}
\theta' =&\ \theta - \beta \nabla_\theta \mathcal{L}_{D_N}(\theta^*) \nonumber \\
       =&\ \theta - \beta \nabla_\theta \mathcal{L}_{D_N}(\theta - \alpha  \nabla_\theta \mathcal{L}_{D_T}(\theta)), \label{maml}
\end{align}
where both $\alpha$ and $\beta$ are hyper-parameters of two-stage learning rate. The above optimization can be conducted with stochastic gradient descent (SGD). In this way, the knowledge learned from $D_T$ can provide a good initial representation that can be effectively fine-tuned by a few examples in $D_N$, and thus achieve fast and robust adaptation.


Note that the technique presented here is a simplified variant of the original MAML, which considers more than just two tasks compared to our case, i.e., a source task ($D_T$) and a target task ($D_N$). If we require to train embeddings for multiple domains simultaneously, we can also extend our approach to deal with multiple $D_T$ and $D_N$.



\section{Experiments}\label{sec:evaluation}
In this section, we present two types of experiments to evaluate the effectiveness of the proposed HiCE model. One is an intrinsic evaluation on a benchmark dataset, and the other is an extrinsic evaluation on two downstream tasks: (1) named entity recognition and (2) part-of-speech tagging.

\subsection{Experimental Settings}
As aforementioned, our approach assumes an initial embedding $T$ trained on an existing corpus $D_T$. As all the baseline models learn embedding from Wikipedia, we train HiCE on WikiText-103~\cite{DBLP:MerityXBS16} with the initial embedding provided by~\citet{DBLP:conf/emnlp/HerbelotB17}\footnote{\url{clic.cimec.unitn.it/~aurelie.herbelot/wiki_all.model.tar.gz}}.

WikiText-103 contains 103 million words extracted from a selected set of articles. From WikiText-103, we select words with an occurrence count larger than 16 as training words. Then, we collect the masked supporting contexts set $S_t$ for each training word $w_t$ with its oracle embedding $T_{w_t}$, and split the collected words into a training set and a validation set. We then train the HiCE model\footnote{\url{github.com/acbull/HiCE}} in the previous introduced episode based $K$-shot learning setting, and select the best hyper-parameters and model using the validation set. 
After we obtain the trained HiCE model, we can either directly use it to infer the embedding vectors for OOV words in new corpus $D_N$, or conduct adaptation on $D_N$ using MAML algorithm as shown in Eq. (\ref{maml}).

\subsection{Baseline Methods}
We compare HiCE with the following baseline models for learning OOV word embeddings.  
\begin{compactitem}
    \item \textbf{Word2Vec:} The local updating algorithm of Word2Vec. The model employs the `Skip-gram' update to learn a new word embedding by predicting its context word vectors. 
    We implement this baseline model with gensim\footnote{\url{radimrehurek.com/gensim/} \label{gensim}}.
    
    \item \textbf{FastText:} FastText is a morphological embedding algorithm that can handle OOV by summing n-gram embeddings. To make fair comparison, we train FastText on WikiText-103, and directly use it to infer the embeddings of OOV words in new datasets. We again use the implementation in gensim$^\text{\ref{gensim}}$.
    
    \item \textbf{Additive:} Additive model~\cite{multimodal} is a purely non-parametric algorithm that averages the word embeddings of the masked supporting contexts $S_t$. Specifically:
    \begin{equation*}
        e^{additive}_t = \frac{1}{|S_t|}\sum\nolimits_{c \in S_t} \frac{1}{|c|}\sum\nolimits_{w \in c}e_w.
    \end{equation*}
    Also, this approach can be augmented by removing the stop words beforehand.
    \item \textbf{nonce2vec:} This algorithm~\cite{DBLP:conf/emnlp/HerbelotB17} is a modification of original gensim Word2Vec implementation, augmented by a better initialization of additive vector, higher learning rates and large context window, etc. We directly used their open-source implementation\footnote{\url{github.com/minimalparts/nonce2vec}}.
    \item $\textbf{\`a \ la\ carte:}$ This algorithm~\cite{DBLP:conf/acl/AroraLMKSS18} is based on an additive model, followed by a linear transformation $A$ that can be learned through an auxiliary regression task. Specifically:
    \begin{equation*}
        e^{\grave{a} \ la\ carte}_t = \frac{A}{|S_t|}\sum\nolimits_{c \in S_t} \sum\nolimits_{w \in c}Ae^{additive}_w
    \end{equation*}
    We conduct experiments by using their open-source implementation\footnote{\url{github.com/NLPrinceton/ALaCarte}}.
\end{compactitem}

\subsection{Intrinsic Evaluation: Evaluate OOV Embeddings on the Chimera Benchmark}

First, we evaluate HiCE on Chimera~\cite{multimodal}, a widely used benchmark dataset for evaluating word embedding for OOV words. 

\paragraph{Dataset} The Chimera dataset simulates the situation when an embedding model faces an OOV word in a real-world application. For each OOV word (denoted as ``chimera''), a few example sentences (2, 4, or 6) are provided. The dataset also provides a set of probing words and the human-annotated similarity between the probing words and the OOV words. To evaluate the performance of a learned embedding, Spearman correlation is used in~\cite{multimodal} to measure the agreement between the human annotations and the machine-generated results.


\begin{table}[t!]
\small
\centering
\begin{tabular}{l|lll}
\toprule
Methods       & 2-shot          & 4-shot          & 6-shot          \\ \midrule
Word2vec                & 0.1459          & 0.2457          & 0.2498          \\
FastText                & 0.1775          & 0.1738          & 0.1294          \\
Additive                & 0.3627          & 0.3701          & 0.3595          \\
Additive, no stop words & 0.3376          & 0.3624          & 0.4080          \\
nonce2vec               & 0.3320          & 0.3668          & 0.3890          \\
$\grave{a} \ la\ carte$      & 0.3634          & 0.3844          & 0.3941          \\ \midrule
HiCE w/o  Morph              & 0.3710          & 0.3872        & 0.4277          \\
HiCE + Morph              & \textbf{0.3796}         & 0.3916         & 0.4253        \\
HiCE + Morph + Fine-tune  & 0.1403          & 0.1837          & 0.3145          \\
HiCE + Morph + MAML       & 0.3781      & \textbf{0.4053}      & \textbf{0.4307}        \\
\midrule
Oracle Embedding& 0.4160 & 0.4381 & 0.4427 \\ \bottomrule
\end{tabular}
\caption{Performance on the Chimera benchmark dataset with different numbers of context sentences, which is measured by Spearman correlation. Baseline results are from the corresponding papers.}
\label{table:bench}
\end{table}

\paragraph{Experimental Results}
Table \ref{table:bench} lists the performance of HiCE and baselines with different numbers of context sentences. In particular, our method (HiCE+Morph+MAML)\footnote{Unless other stated, HiCE refers to HiCE + Morph + MAML.} achieves the best performance among all the other baseline methods under most settings. Compared with the current state-of-the-art method, $\grave{a}  \ la\ carte$, the relative improvements (i.e., the performance difference divided by the baseline performance) of HiCE are 4.0\%, 5.4\% and  9.3\% in terms of 2,4,6-shot learning, respectively.   
We also compare our results with that of the oracle embedding, which is the embeddings trained from $D_T$, and used as ground-truth to train HiCE. This results can be regarded as an upper bound. As is shown, when the number of context sentences (K) is relatively large~(i.e., $K=6$), the performance of HiCE is on a par with the upper bound (Oracle Embedding) and the relative performance difference is merely 2.7\%. This indicates the significance of using an advanced aggregation model.

Furthermore, we conduct an ablation study to analyze the effect of morphological features. By comparing HiCE with and without Morph, we can see that morphological features are helpful when the number of context sentences is relatively small~(i.e., 2 and 4 shot). This is because morphological information does not rely on context sentences, and can give a good estimation when contexts are limited. However, in 6-shot setting, their performance does not differ significantly. 

In addition, we analyze the effect of MAML by comparing HiCE with and without MAML. We can see that adapting with MAML can improve the performance when the number of context sentences is relatively large~(i.e., 4 and 6 shot), as it can mitigate the semantic gap between source corpus $D_T$ and target corpus $D_N$, which makes the model better capture the context semantics in the target corpus. Also we evaluate the effect of MAML by comparing it with fine-tuning. The results show that directly fine-tuning on target corpus can lead to extremely bad performance, due to the insufficiency of data. On the contrary, adapting with MAML can leverage the source corpus's information as regularization to avoid over-fitting.

\begin{table*}[ht]
\centering
\begin{tabular}{l||c|c|c|l|l|l}
\toprule
\multirow{2}{*}{Methods} & \multicolumn{2}{c|}{Named Entity Recognition (F1-score)}              & \multicolumn{4}{c}{POS Tagging (Acc)} \\\cline{2-7} 
                      & Rare-NER & Bio-NER &  \multicolumn{4}{c}{Twitter POS}               \\\midrule
Word2vec               & 0.1862                         & 0.7205                            & \multicolumn{4}{c}{0.7649}                            \\
FastText             & 0.1981                           & 0.7241                            & \multicolumn{4}{c}{0.8116}                            \\
Additive              & 0.2021                         & 0.7034                            & \multicolumn{4}{c}{0.7576}                            \\
nonce2vec             & 0.2096                           & 0.7289                            & \multicolumn{4}{c}{0.7734}                            \\
$\grave{a} \ la\ carte$             & 0.2153                           & 0.7423                            & \multicolumn{4}{c}{0.7883}                            \\ \midrule
HiCE w/o Morph          & 0.2394                           & 0.7486                            &\multicolumn{4}{c}{0.8194}                            \\
HiCE + Morph          & 0.2375                           & 0.7522                            & \multicolumn{4}{c}{0.8227}                            \\
HiCE + Morph + MAML   & \textbf{0.2419}                           & \textbf{0.7636}                            & \multicolumn{4}{c}{\textbf{0.8286}}   \\ \bottomrule
\end{tabular}

\caption{Performance on Named Entity Recognition and Part-of-Speech Tagging tasks. All methods are evaluated on test data containing OOV words. Results demonstrate that the proposed approach, HiCE + Morph + MAML, improves the downstream model by learning better representations for OOV words.  }
\label{table:down}
\end{table*}

\subsection{Extrinsic Evaluation: Evaluate OOV Embeddings on Downstream Tasks}
To illustrate the effectiveness of our proposed method in dealing with OOV words, we evaluate the resulted embedding on two downstream tasks: (1)  named entity recognition (NER) and (2) part-of-speech (POS) tagging.

\paragraph{Named Entity Recognition}
NER is a semantic task with a goal to extract named entities from a sentence. Recent approaches for NER take word embedding as input and leverage its semantic information to annotate named entities. Therefore, a high-quality word embedding has a great impact on the NER system. We consider the following two corpora, which contain abundant OOV words, to mimic the real situation of OOV problems. 


\begin{compactitem}
    \item {\bf Rare-NER}: This NER dataset~\cite{DBLP:conf/aclnut/DerczynskiNEL17} focus on unusual, previously-unseen entities in the context of emerging discussions, which are mostly OOV words. 

    \item {\bf Bio-NER}: The JNLPBA 2004 Bio-entity recognition dataset~\cite{DBLP:conf/bionlp/CollierK04} focuses on technical terms in the biology domain, which also contain many OOV words. 
\end{compactitem}
Both datasets use entity-level F1-score as an evaluation metric.
We use the WikiText-103 as $D_T$, and these datasets as $D_N$. We select all the OOV words in the dataset and extract their context sentences. Then, we train different versions of OOV embeddigns based on the proposed approaches and the baseline models. Finally, the inferred embedding is used in an NER system based on the Bi-LSTM-CRF~\cite{DBLP:conf/naacl/LampleBSKD16} architecture to predict named entities on the test set. We posit a higher-quality OOV embedding results in better downstream task performance. 

As we mainly focus on the quality of OOV word embeddings, we construct the test set by selecting sentences which have at least one OOV word. In this way, the test performance will largely depend on the quality of the OOV word embeddings. After the pre-processing, Rare-NER dataset contains 6,445 OOV words and 247 test sentences, while Bio-NER contains 11,748 OOV words and 2,181 test sentences. Therefore, Rare-NER has a high ratio of OOV words per sentence. 

\paragraph{Part-of-Speech Tagging}
Besides NER, we evaluate the syntactic information encoded in HiCE through a lens of part-of-speech (POS) tagging, which is a standard task with a goal to identify which grammatical group a word belongs to. We consider the Twitter social media POS dataset~\cite{D11-1141}, which contains many OOV entities. The dataset is comprised of 15,971 English sentences collected from Twitter in 2011. Each token is tagged manually into 48 grammatical groups, consisting of Penn Tree Bank Tag set and several Twitter-specific classes. The performance of a tagging system is measured by accuracy. 
Similar to the previous setting, we use different updating algorithms to learn the embedding of OOV words in this dataset, and show different test accuracy results given by learned Bi-LSTM-CRF tagger. The dataset contains 1,256 OOV words and 282 test sentences.

\begin{figure*}[t]
\begin{center}
\includegraphics[width=2\columnwidth]{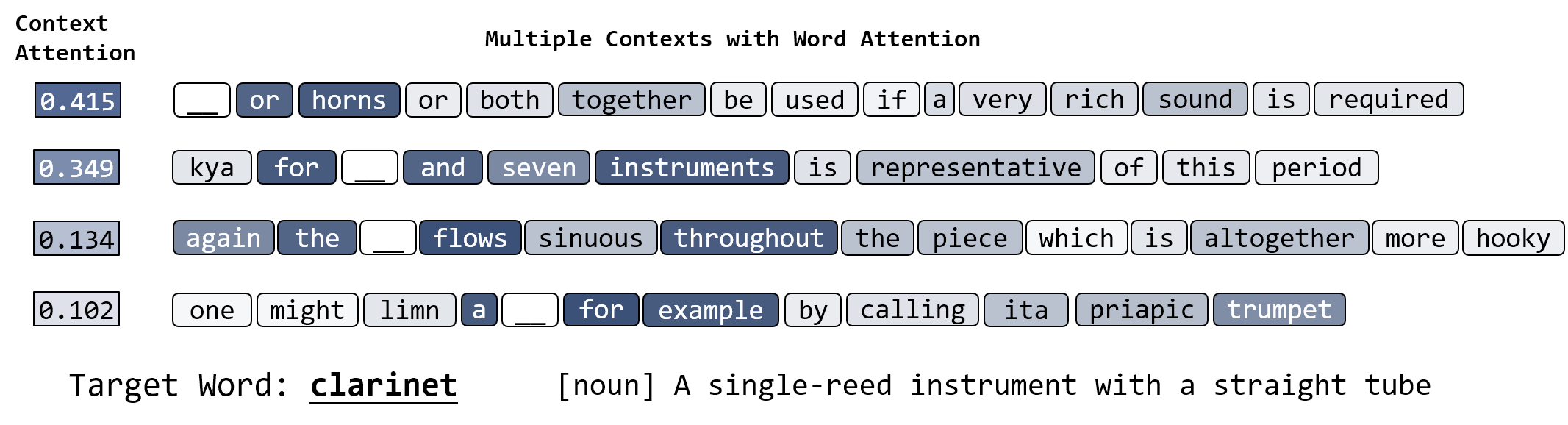}
\caption{Visualization of attention distribution over words and contexts. }\label{fig:att1}
\end{center}
\end{figure*}

\begin{table*}[ht!]
\centering
\small

\begin{tabular}{l|l|l|l}
\toprule
\textbf{OOV Word}       & \textbf{Contexts}    & \textbf{Methods}  & \textbf{Top-5 similar words (via cosine similarity)}                                \\ \midrule
\multirowcell{3}{\textbf{scooter}} & \multirowcell{3}[0ex][l]{We all need vehicles like bmw\\ c1 \underline{scooter} that allow more social\\ interaction while using them~...}  & Additive & the, and, to, of, which                     \\
                    &    & FastText & cooter, pooter, footer, soter, sharpshooter                 \\
                    &    & \textbf{HiCE}     & cars, motorhomes, bmw, motorcoaches, microbus
\\ \midrule
\multirowcell{3}{\textbf{cello}}  & \multirowcell{3}[0ex][l]{The instruments I am going to\\ play in the band service are\\ the euphonium and the \underline{cello}~...} & Additive & the, and, to, of, in                                \\ 
                    &    & FastText & celli, cellos, ndegéocello, cellini, cella          \\ 
                    &    & \textbf{HiCE}     & piano, orchestral, clarinet, virtuoso, violin       \\ \midrule
\multirowcell{3}{\textbf{potato}} & \multirowcell{3}[0ex][l]{It started with a green salad\\ followed by a mixed grill with\\ rice chips \underline{potato}~...} & Additive & and, cocoyam, the, lychees, sapota                  \\ 
                    &    & FastText & patatoes, potamon, potash, potw, pozzato            \\
                    &    & \textbf{HiCE}     & vegetables, cocoyam, potatoes, calamansi, sweetcorn \\ 
\bottomrule
\end{tabular}
\caption{For each OOV in Chimera benchmark, infer its embedding using different methods, then show top-5 words with similar embedding to the inferred embedding. HiCE can find words with most similar semantics. }\label{tab:sim}
\end{table*}

\paragraph{Results}
Table \ref{table:down} illustrates the results evaluated on the downstream tasks. HiCE outperforms the baselines in all the settings. Compared to the best baseline $\grave{a} \ la\ carte$, the relative improvements are 12.4\%, 2.9\% and 5.1\% for Rare-NER, Bio-NER, and Twitter POS, respectively. As aforementioned, the ratio of OOV words in Rare-NER is high. As a result, all the systems perform worse on Rare-NER than Bio-NER, while HiCE reaches the largest improvement than all the other baselines. Besides, our baseline embedding is trained on Wikipedia corpus (WikiText-103), which is quite different from the bio-medical texts and social media domain.
The experiment demonstrates that HiCE trained on $D_T$ is already able to leverage the general language knowledge which can be transferred through different domains, and adaptation with MAML can further reduce the domain gap and enhance the performance.

\subsection{Qualitative Evaluation of HiCE} 
To illustrate how does HiCE extract and aggregate information from multiple context sentences, we visualize the attention weights over words and contexts. 
We demonstrate an example in Figure \ref{fig:att1}, where we choose four sentences in chimera dataset, with ``clarinet'' (a woodwind instrument) as the OOV word. From the attention weight over words, we can see that the HiCE puts high attention on words that are related to instruments, such as ``horns'', ``instruments'', ``flows'', etc. From the attention weight over contexts, we can see that HiCE assigns the fourth sentence the lowest context attention, in which the instrument-related word ``trumpet'' is distant from the target placeholder, making it harder to infer the meaning by this context. This shows HiCE indeed distinguishes important words and contexts from the uninformative ones. 

Furthermore, we conduct a case study to show how well the inferred embedding for OOV words capture their semantic meaning. We randomly pick three OOV words with 6 context sentences in Chimera benchmark, use additive, fastText and HiCE to infer the embeddings. Next, we find the top-5 similar words with the highest cosine similarity. As is shown in Table \ref{tab:sim}, Additive method can only get embedding near to neutral words as ``the'', ``and'', etc, but cannot capture the specific semantic of different words. FastText can find words with similar subwords, but representing totally different meaning. For example, for OOV ``scooter'' (a motor vehicle), FastText finds ``cooter'' as the most similar word, which looks similar in character-level, but means a river turtle actually. Our proposed HiCE however, can capture the true semantic meaning of the OOV words. For example, it finds ``cars'', ``motorhomes'' (all are vehicles) for ``scooter'', and finds ``piano'', ``orchestral'' (all are instruments) for ``cello'', etc. This case study shows that HiCE can truly infer a high-quality embedding for OOV words.
\section{Related Work}\label{sec:relate}

\paragraph{OOV Word Embedding}
Previous studies of handling OOV words were mainly based on two types of information: 1) context information and 2) morphology features.

The first family of approaches follows the distributional hypothesis~\cite{firth1957synopsis} to infer the meaning of a target word based on its context. If sufficient observations are given, simply applying existing word embedding techniques (e.g., word2vec) can already learn to embed OOV words. However, in a real scenario, mostly the OOV word only occur for a very limited times in the new corpus, which hinders the quality of the updated embedding~\cite{multimodal, DBLP:conf/emnlp/HerbelotB17}. Several alternatives have been proposed in the literature. \citet{multimodal} proposed additive method by using the average embeddings of context words~\cite{multimodal} as the embedding of the target word. \citet{DBLP:conf/emnlp/HerbelotB17} extended the skip-gram model to \textit{nonce2vec} by initialized with additive embedding, higher learning rate and window size. \citet{DBLP:conf/acl/AroraLMKSS18} introduced $\textit{\`a \ la\ carte}$, which augments the additive method by a linear transformation of context embedding. 

The second family of approaches utilizes the morphology of words (e.g., morphemes, character n-grams and character) to construct embedding vectors of unseen words based on sub-word information. For example, \citet{DBLP:conf/conll/LuongSM13} proposed a morphology-aware word embedding technique by processing a sequence of morphemes with a recurrent neural network. \citet{bojanowski2016enriching} extended skip-gram model by assigning embedding vectors to every character n-grams and represented each word as the sum of its n-grams. \citet{DBLP:conf/emnlp/PinterGE17} proposed MIMICK to induce word embedding from character features with a bi-LSTM model. Although these approaches demonstrate reasonable performance, they rely mainly on morphology structure and cannot handle some special type of words, such as transliteration, entity names, or technical terms.

Our approach utilizes both pieces of information for an accurate estimation of OOV embeddings. To leverage limited context information, we apply a complex model in contrast to the linear transformation used in the past, and learn to embed in a few-shot setting. We also show that incorporating morphological features can further enhance the model when the context is extremely limited (i.e., only two or four sentences).

    

\paragraph{Few-shot learning} 

The paradigm of learning new tasks from a few labelled observations, referred to as few-shot learning, has received significant attention. The early studies attempt to transfer knowledge learned from tasks with sufficient training data to new tasks. They mainly follow a pre-train then fine-tune paradigm~\cite{DBLP:conf/icml/DonahueJVHZTD14, DBLP:journals/jmlr/Bengio12, DBLP:conf/emnlp/ZophYMK16}. Recently, meta-learning is proposed and it achieves great performance on various few-shot learning tasks. The intuition of meta-learning is to learn generic knowledge on a variety of learning tasks, such that the model can be adapted to learn a new task with only a few training samples. 
Approaches for meta-learning can be categorized by the type of knowledge they learn. 
    (1) Learn a metric function that embeds data in the same class closer to each other, including Matching Networks~\cite{DBLP:conf/nips/VinyalsBLKW16}, and Prototypical Networks~\cite{DBLP:conf/nips/SnellSZ17}. The nature of metric learning makes it specified on classification problems.
    (2) Learn a learning policy that can fast adapt to new concepts, including a better weight initialization as MAML~\cite{DBLP:conf/icml/FinnAL17} and a better optimizer~\cite{OPTIMIZATION}. This line of research is more general and can be applied to different learning paradigms, including both classification and regression.

There have been emerging research studies that utilize the above meta-learning algorithms to NLP tasks, including language modelling~\cite{DBLP:conf/nips/VinyalsBLKW16}, text classification~\cite{DBLP:conf/naacl/YuGYCPCTWZ18}, machine translation~\cite{DBLP:conf/emnlp/GuWCLC18}, and relation learning~\cite{DBLP:conf/emnlp/XiongYCGW18, DBLP:conf/aaai/Han0S18}.
In this paper, we propose to formulate the OOV word representation learning as a few-shot regression problem. We first show that pre-training on a given corpus can somehow solve the problem. To further mitigate the semantic gap between the given corpus with a new corpus, we adopt model-agnostic meta-learning (MAML)~\cite{DBLP:conf/icml/FinnAL17} to fast adapt the pre-trained model to new corpus.

\paragraph{Contextualized Embedding}
The HiCE architecture is related to contextualized representation learning~\cite{DBLP:conf/naacl/PetersNIGCLZ18,bert}. However, their goal is to get a contextualized embedding based on a given sentence, with word or sub-word embeddings as input. In contrast, our work utilizes multiple contexts to learn OOV embeddings. This research direction is orthogonal to their goal. In addition, the OOV embeddings learned by ours can be served as inputs to ELMO and BERT, helping them to deal with OOV words.


\section{Conclusion}\label{sec:conclusion}We studied the problem of learning accurate embedding for Out-Of-Vocabulary word and augment them to a per-trained embedding by only a few observations. We formulated the problem as a K-shot regression problem and proposed a hierarchical context encoder (HiCE) architecture that learns to predict the oracle OOV embedding by aggregating only K contexts and morphological features.  We further adopt MAML for fast and robust adaptation to mitigate semantic gap between corpus. Experiments on both benchmark corpus and downstream tasks demonstrate the superiority of HiCE over existing approaches.




\section*{Acknowledgments}\label{sec:acknowledgments}
This work is partially supported by NSF RI-1760523, NSF III-1705169, NSF CAREER Award 1741634, and Amazon Research Award.
\bibliography{acl}
\bibliographystyle{acl_natbib}
\end{document}